# CEMSSL: Conditional Embodied Self-Supervised Learning is All You Need for High-precision Multi-solution Inverse Kinematics of Robot Arms

Qu Weiming[†]
*School of Intelligence Science and Technology*
*Peking University*
Beijing, China
quweiming@pku.edu.cn

Liu Tianlin[†]
*School of Intelligence Science and Technology*
*Peking University*
Beijing, China
liutl@pku.edu.cn

Du Jiawei
*School of Intelligence Science and Technology*
*Peking University*
Beijing, China
neodydu@stu.pku.edu.cn

Luo Dingsheng*, Member, IEEE
*School of Intelligence Science and Technology*
*Peking University*
Beijing, China
dsluo@pku.edu.cn

*Abstract*—In the field of signal processing for robotics, the inverse kinematics of robot arms presents a significant challenge due to multiple solutions caused by redundant degrees of freedom (DOFs). Precision is also a crucial performance indicator for robot arms. Current methods typically rely on conditional deep generative models (CDGMs), which often fall short in precision. In this paper, we propose Conditional Embodied Self-Supervised Learning (CEMSSL) and introduce a unified framework based on CEMSSL for high-precision multi-solution inverse kinematics learning. This framework enhances the precision of existing CDGMs by up to 2-3 orders of magnitude while maintaining their original properties. Furthermore, our method is extendable to other fields of signal processing where obtaining multi-solution data in advance is challenging, as well as to other problems involving multi-solution inverse processes.

## I. Introduction

In the realm of signal processing, multi-solution problems are widespread, including tasks such as speech synthesis [1-3], image inpainting [4-6], and trajectory prediction [7-10]. Inverse kinematics of robot arms is another typical example. Multi-solution inverse kinematics primarily arise from the redundant degrees of freedom (DOFs) in robot arms. This redundancy allows for multiple joint configurations to achieve the same end-effector position, offering benefits such as increased flexibility and robustness, but also posing challenges in inverse kinematic model learning. The ability to find and process multiple solutions significantly enhances the success rate of trajectory planning in environments with obstacles (as shown in Fig.1) and ensures the robot arm remains functional despite partial joint damage.

Traditional methods for multi-solution inverse kinematic model learning involve dividing the entire joint space into multiple subspaces, each with a single solution, and then combining these solutions [11-13]. However, this often requires an independent analysis for each problem, and can be difficult or even infeasible for high spatial dimensions. In recent years, deep neural networks have enabled methods based on conditional deep generative models, such as conditional variational autoencoder (CVAE) [14,15], conditional generative adversarial network (CGAN) [16-18], and conditional invertible neural network (CINN) [19,20], to directly model the distribution of multiple solutions. However, these methods require labeled datasets for training, and the learned multi-solution distribution depends heavily on the distribution of the collected data. Moreover, these methods tend to focus more on the distribution of multiple solutions, rather than the precision of the inverse model.

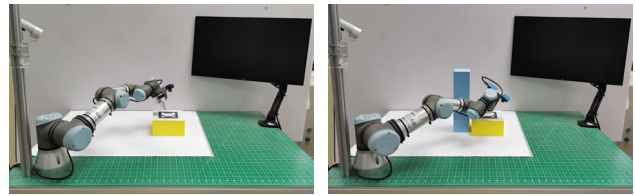

Fig. 1. The impact of multiple solutions in robot arm trajectory planning. In the presence of obstacles, the solution on the left results in a collision, whereas the solution on the right avoids a collision.

To address the aforementioned shortcomings, in this paper, we investigate high-precision multi-solution inverse kinematic model learning for robot arms from a generative model perspective. Specifically, we propose conditional embodied self-supervised learning (CEMSSL) to address the problem that the original embodied self-supervised learning (EMSSL) model can only learn one particular solution among multiple solutions and cannot obtain multiple solutions [9]. We also adopt model ensemble to further enhance the multi-solution learning capability of CEMSSL. Additionally, we present a unified framework for robot arm high-precision multi-solution inverse model learning, which is also applicable to other CDGMs. While CEMSSL is a general framework that is applicable to any signal processing tasks involving multi-solution inverse processes, we demonstrate its utility in the application of robot arm inverse kinematic model learning. In summary, the main contributions of this paper are summarized as follows:

- We propose conditional embodied self-supervised learning (CEMSSL) for robot arm multi-solution inverse kinematics.
- We present a unified framework for robot arm high-precision multi-solution inverse kinematics, which can significantly improve the precision of CDGMs while maintaining their original properties.
- Our method is broadly applicable to various signal processing problems, where obtaining multi-solution data in advance is challenging.

## II. PROBLEM FORMULATION

For the multi-solution inverse kinematic model learning of robot arms, the generated joint angles must meet certain constraints. Specifically, the end position corresponding to the generated joint angles must match the desired end position, which serves as the label. Therefore, the multi-solution learning problem can be viewed as a conditional generation problem. In this context, for the joint angles $q$, the end position $p$, and the hidden variable $z$, our goal is to capture their joint probability distribution $P(p, q, z)$

Given independently and identically distributed (iid) training data $\{(q^{(i)}, p^{(i)}, z^{(i)})\}_{i=1}^{N}$, the objective of multi-solution inverse model learning is to find the optimal parameters $\theta^*$ such that:

$$\theta^* = \arg\max_{\theta} \sum_{i=1}^{N} \log P\left(p^{(i)} \mid FM\left(IM_\theta(p^{(i)}, z^{(i)})\right)\right) \quad (1)$$

Where $z^{(i)} \sim N(0, I), i = 1, 2, \ldots, N$, FM and IM represent the forward model and inverse model.

## III. METHODOLOGY

In the original EMSSL method [21], the inverse model can only learn a particular solution among multiple solutions, but not the distribution of multiple solutions. In this section, we first propose conditional embodied self-supervised learning (CEMSSL) to solve this issue, we then introduce model ensemble to deal with the mode collapse problem, and finally we present a CEMSSL-based unified framework for high-precision multi-solution inverse kinematics model learning, which is applicable to other CDGMs.

### A. Conditional Embodied Self-Supervised Learning

In CEMSSL, the input of the inverse model has an additional hidden variable $z$, which is sampled from the Gaussian distribution $N(0, I)$, and the input of the original inverse model, i.e., the desired position of the end of the robot arm, is used as the conditional input. In contrast to other CDGMs, the dataset for our method is obtained by iteratively coordinating sampling and training within CEMSSL. The CEMSSL algorithm is shown in Algorithm1.

**Algorithm 1** Conditional Embodied Self-Supervised Learning

**Input:** Forward Model FM, Inverse Model IM, Unlabeled dataset of states and relative positions of the robot arm $U = \{p^{(i)}\}_{i=1}^{N}$, Maximum number of iterations $T$, Epochs $E$, Number of small batch samples for inference $M_R$, Number of small batch samples for training $M_T$, Number of parallel computation threads $K$, Learning rate $\eta$

1: Initialize the parameters $\theta$ of the inverse model IM randomly
   Initialize the sample dataset: $D \leftarrow \emptyset$
   Initialize the joint angle variation dataset of IM inference: $Q \leftarrow \emptyset$
   Initialize the relative position dataset of FM computation: $P \leftarrow \emptyset$
   Initialize the random noise dataset: $Z \leftarrow \emptyset$
2: **for** $t = 1 \ldots T$ **do**
3:   **\*\* Data Sampling \*\***
4:   Empty the dataset $D, Q, P, Z$
5:   **for** $n = 1, 2 \ldots N_{BU}$ **do**
6:     Sample small batches from $U$: $B \leftarrow \{p^{(m)}\}_{m=1}^{M_R}$
7:     Sampling a mini-batch of samples $Z_n$ from the Gaussian distribution N(0, I) : $Z_n \leftarrow \{z^{(m)}\}_{m=1}^{M_R}$
8:     Update the dataset: $Q \leftarrow Q \cup Q_n, Z \leftarrow Z \cup Z_n$
9:   **end for**
10:   $j \leftarrow 0$
11:   **repeat**
12:     According to the number of threads $K$, pop the joint angle: $\{q_{1+j}, q_{2+j}, \ldots, q_{K+j}\}$
13:     FM parallel computation: $P_j = FM(S_j, Q_j)$
14:     Update the dataset: $P \leftarrow P \cup P_j$
15:     $j = j + 1$
16:   **until** the data in $Q$ have been computed
17:   Update the sample dataset: $D \leftarrow \{Z, Q, P\}$
18:   **\*\* Model Training \*\***
19:   **for** $e = 1 \ldots E$ **do**
20:     **for** $n = 1, 2 \ldots N_{BD}$ **do**
21:       Sample small batches from $D$: $B \leftarrow \{(s^{(m)}, \Delta q^{(m)}, \Delta p^{(m)})\}_{m=1}^{M_T}$
22:       $L(\theta) = \frac{1}{M_T}\sum_{i=1}^{M_T}\left(\Delta q^{(m)} - IM((s^{(m)}, \Delta p^{(m)}))\right)^2$
23:       Update $\theta$ with GD: $\theta \leftarrow \theta - \eta \nabla_\theta L(\theta)$
24:     **end for**
25:   **end for**
26: **end for**

In the data sampling phase, given the desired end position of the robot arm, the hidden variable $z$ is sampled from the Gaussian distribution $N(0, I)$. The desired end position of the robot arm and the hidden variable $z$ are input to the inverse model to infer the corresponding joint angle. The forward model is then used to predict the end position of the joint angle.

In the model training phase, the previous predicted end position of the robot arm and the hidden variable $z$ are input to the same inverse model to infer a new joint angle. The joint angle obtained from the data sampling phase is used to perform supervised learning of the new joint angle. The loss function for the model training phase is as follows:

$$L(\theta) = \frac{1}{N}\sum_{i=1}^{N}\left(q^{(i)} - IM(p^{(i)}, z^{(i)})\right)^2 \quad (2)$$

### B. Model Ensemble

A common problem with learning distributions is that learning becomes difficult when the target distribution is multimodal, such as the mode collapse problem, i.e., only one mode of the multimodal target distribution is learned, which leads to a series of homogeneous results [22]. This problem also exists for the multi-solution inverse kinematic model learning of the robot arm.

To overcome this problem, we adopt model ensemble [23,24] to train several initialized inverse models of different networks simultaneously. Using different networks can better cover the distribution of multiple solutions than using a single network.

*C. CEMMSL-Based Unified Framework*

Due to the similarity in the input and output between the decoder of CVAE, the generator of CGAN, the inverse process in CINN, and the inverse model in CEMSSL (i.e., the input is the desired position of the end of the robot arm and the hidden variable $z$, and the output is the joint angle), they can be unified and used for online iterative learning [18]. As shown in Fig.2, the inverse model is first pretrained by CDGMs. Then, CEMSSL is employed to perform online iterative learning based on the pretrained inverse model. For CVAE, CGAN, and CINN, the purpose of online iterative learning is to use the forward model to guide the update of the inverse model based on the original pretrained inverse model. During this process, the online iterative learning serves as a fine-tuning step, maintaining the properties of original models. Hence, the proposed framework can achieve high-precision multi-solution inverse kinematic models while maintaining the model properties of CVAE, CGAN, and CINN.

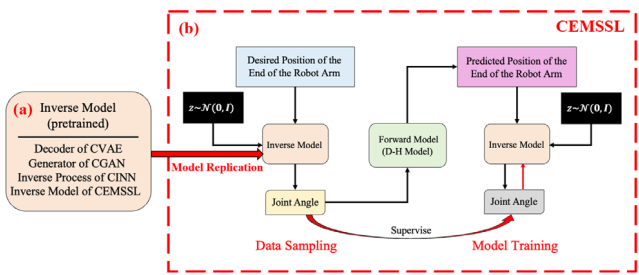

Fig. 2. The architecture of CEMSSL-base unified framework for high-precision multi-solution inverse kinematics model learning.

## IV. EXPERIMENTS

*A. Environment Setup*

To evaluate the proposed method, we set up a series of experiments on a real UR3 robot arm platform, as well as a simulation platform in gazebo. The real UR3 robot arm platform is shown in Fig.3. Our tasks focus solely on the end-effector position, disregarding its orientation, and hence the UR3 robot arm with 6 DOFs exhibits a redundancy of 3 DOFs in this context.

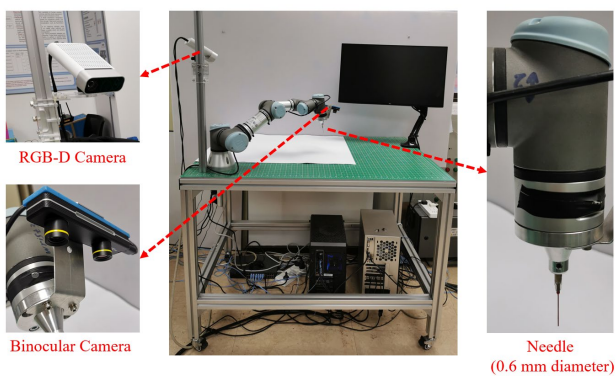

Fig. 3. The real UR3 robot arm platform for the experiments.

*B. Implementation Details*

We test various hidden variable dimensions, including 1, 2, 4, 6, and 8. The inverse kinematic model is failed to converge when the hidden variable dimension is 1, but for other dimensions, the inverse model converges to 0.02 mm. We observe that when the dimension is too low or too high, the diversity of the learned multiple solutions is relatively poor. In our experiments, we determine that a dimension of 6 is most suitable.

Regarding the number of model integrations, we test five different settings: 1, 2, 4, 6, and 8. We observe that with too few ensembles, the distribution of multiple solutions could only cover part of the modes. However, with a sufficient number of ensembles (about 6), all modes could be adequately covered. The hyperparameters are shown in Table I.

**Metrics:** We evaluate the model performance in terms of the diversity of solutions, i.e., the ability to learn multiple modalities of multiple solutions, as well as the precision of the inverse kinematic model. Mathematically,

$$precison = \sum_{i=1}^{N} \left\| \boldsymbol{p}^{(i)} - FM\left(IM_{\hat{\boldsymbol{\theta}}}(\boldsymbol{p}^{(i)}, \boldsymbol{z}^{(i)})\right) \right\|_2 \quad (3)$$

**Baselines:** We compare our proposed method against three published CDGM baselines briefly described below.

CVAE [4]: Sohn et al. propose to condition the latent representation to generate diverse and realistic data.

CGAN [12]: Mirza et al. add a conditional input to both the generator and the discriminator, allowing CGAN to generate data with specific conditions.

CINN [8]: Ardizzone et al. directly map the input into the hidden variable for a given condition, trained with likelihood loss (LHL).

TABLE I. HYPERPARAMETERS SETTINGS

| Hyperparameters | Value |
| --- | --- |
| Activation | ReLU (Hidden) Sigmoid (Output) |
| Optimizer | Adam |
| Learning Rate | 0.0015 |
| Batch Size (Inference) | 512 |
| Batch Size (Training) | 128 |
| Parallel Computation Threads | 6 |
| Maximum Iteration Number | 200 |
| Epoch | 10 |
| Hidden Variable Dimension | 6 |
| Number of Model Ensemble | 6 |
| Number of Network Layers | 6 |
| Network Type | FCNN |
| Network Structure | (3, Zdim) → 1024 → 512 →256 → 128 → 6 |

*C. Quantitative Results and Analysis*

The precision of different multi-solution inverse model learning methods is shown in Table II. From the results, we observe that the CEMSSL-based method performs the best, while the CVAE-based method performs the worst.

One reason for the poor performance of CVAE may be the use of two losses during training: MSE loss and KL divergence loss, which preset a trade-off. If the MSE loss is heavily weighted, the precision may be relatively high, but the distribution of the hidden variables may deviate significantly from the standard normal distribution. Conversely, if the KL divergence loss is heavily weighted, the distribution of the hidden variables may be closer to the standard normal distribution, but the precision may suffer.

To validate the effectiveness of the unified framework for other CDGMs, we use CEMSSL to improve CVAE, CGAN,

and CINN models based on the previous training results. The experimental results, presented in Fig.4, demonstrate that the framework can significantly enhance the precision of the inverse kinematic models based on other CDGMs by approximately 2-3 orders of magnitude.

TABLE II. PRECISION COMPARISON OF DIFFERENT METHODS

| Method | Precision (mm) |
|--------|----------------|
| CVAE   | 11.45          |
| CGAN   | 4.78           |
| CINN   | 1.73           |
| CEMSSL | 0.02           |

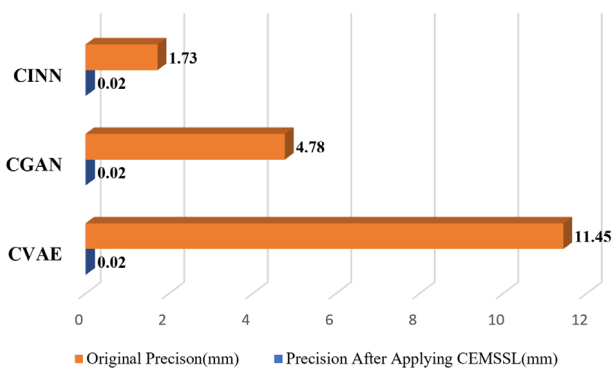

Fig. 4. Comparison of the performance before and after applying the CEMSSL-based framework.

### D. Qualitative Results

To evaluate the diversity, we conduct experiments on the same target position and select four typical patterns for each method. For a given target position, the multi-solution distribution can be intuitively divided into four modes: front, back, left, and right. Hence, as shown in Fig.5, we evaluate the learning of the multi-solution distribution based on whether the method is able to capture these four modes.

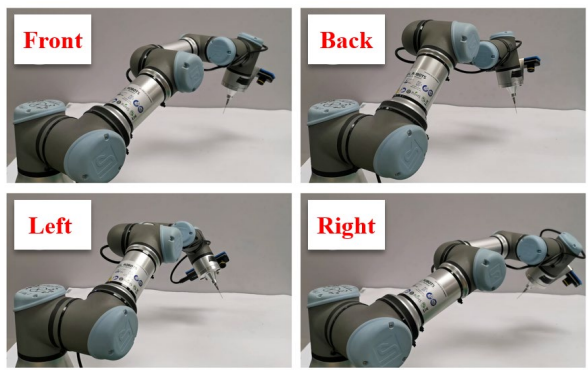

Fig. 5. Samples of the multiple solutions based on CEMSSL for the same target position.

As for the evaluation of the precision, the experimental task involves manipulating the robot arm with the learned inverse kinematic model to position the needle at the end of the robot arm at the intersection of two straight lines located at the center of a rectangular block. The samples of the high-precision reaching multi-solution results are shown in Fig.6.

Fig.7 shows a comparison of the performance before and after the improvement of CEMSSL-based framework. The black dot represents the target point. With CVAE only, the end of the robot arm deviated from the target point, while the end of the robot arm approached the target point with high precision after applying CEMSSL. Notably, the overall shape of the robot arm is only slightly adjusted, indicating that our proposed framework enhances precision while maintaining the properties of the original model. Similar results can also be observed for CGAN and CINN.

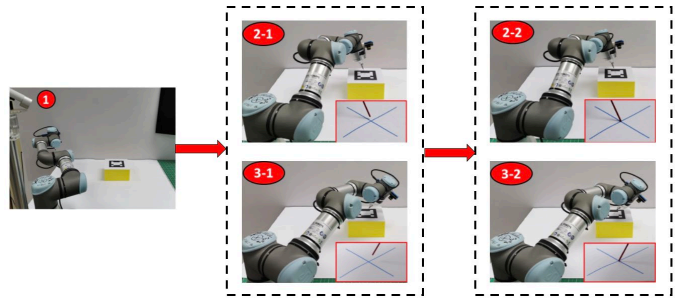

Fig. 6. Samples of high-precision reaching multi-solutions results based on CEMSSL.

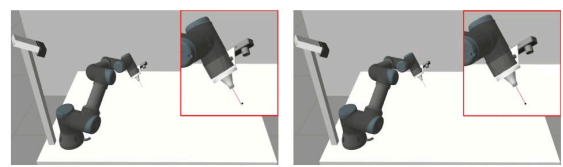

(a) CVAE        (b) CVAE+CEMSSL

Fig. 7. Comparison of the performance before and after the improvement.

## V. CONCLUSION AND BROADER IMPACT

In this paper, we propose conditional embodied self-supervised learning (CEMSSL) for multi-solution inverse kinematics of robot arms, which is a significant challenge in signal processing for robotics. CEMSSL introduces hidden variables and uses the inverse kinematic model input as a condition, effectively addressing the limitations of EMSSL. Model ensemble is adopted to further enhance the learning capability. Experimental results demonstrate that CEMSSL can accurately learn the multi-solution distribution of the robot arm with extensive coverage. Additionally, we present a unified framework for high-precision multi-solution inverse kinematics model learning based on CMESSL, which is also applicable to other conditional deep generative models (CDGMs). This framework significantly improves the precision by approximately 2-3 orders of magnitude.

Since CEMSSL is able to autonomously explore data to learn multi-solution distributions, our method, although exemplified through robot arm inverse kinematics learning, can be extended to other signal processing research areas where obtaining multi-solution data in advance is challenging, or to other problems involving multi-solution inverse processes.

ACKNOWLEDGMENT

This work is supported in part by the National Natural Science Foundation of China (No. 62176004, No. U1713217), Intelligent Robotics and Autonomous Vehicle Lab (RAV), Wuhan East Lake High-Tech Development Zone National Comprehensive Experimental Base of Governance of Intelligent Society, and High-performance Computing Platform of Peking University.

The authors would also like to thank the anonymous reviewers for their attention and valuable suggestions.

## REFERENCES

[1] Guo X, Hu W, Ni C, et al. "Blind Inpainting with Object-Aware Discrimination for Artificial Marker Removal". In: IEEE International Conference on Acoustics, Speech and Signal Processing (ICASSP). 2024: 1516-1520.

[2] Yu L, Gao Y, Pakdaman F, et al. "Panoramic Image Inpainting with Gated Convolution and Contextual Reconstruction Loss". In: IEEE International Conference on Acoustics, Speech and Signal Processing (ICASSP). 2024: 4255-4259.

[3] Hong S, Choi K. "Correcting Faulty Road Maps by Image Inpainting". In: IEEE International Conference on Acoustics, Speech and Signal Processing (ICASSP). 2024: 2710-2714.

[4] Maiti S, Peng Y, Choi S, et al. "VoxtLM: Unified Decoder-Only Models for Consolidating Speech Recognition, Synthesis and Speech, Text Continuation Tasks". In: IEEE International Conference on Acoustics, Speech and Signal Processing (ICASSP). 2024: 13326-13330.

[5] Chu K, Collins L, Mainsah B. "Using automatic speech recognition and speech synthesis to improve the intelligibility of cochlear implant users in reverberant listening environments". In: IEEE International Conference on Acoustics, Speech and Signal Processing (ICASSP). 2020: 6929-6933.

[6] Rosenbaum T, Cohen I, Winebrand E. "Attenuation of Acoustic Early Reflections In Television Studios Using Pretrained Speech Synthesis Neural Network". In: IEEE International Conference on Acoustics, Speech and Signal Processing (ICASSP). 2022: 7422-7426.

[7] Li J, Ma H, Tomizuka M. "Conditional generative neural system for probabilistic trajectory prediction". In: IEEE/RSJ International Conference on Intelligent Robots and Systems (IROS). 2019: 6150-6156.

[8] Liu Y, Zhang J, Fang L, et al. "Multimodal motion prediction with stacked transformers" In: Proceedings of the IEEE/CVF Conference on Computer Vision and Pattern Recognition (CVPR). 2021: 7577-7586.

[9] Korbmacher R, Tordeux A. "Review of pedestrian trajectory prediction methods: Comparing deep learning and knowledge-based approaches". IEEE Transactions on Intelligent Transportation Systems, 2022, 23(12): 24126-24144.

[10] Chen J, Wang W, Chen J, et al. "Dynamic vehicle graph interaction for trajectory prediction based on video signals". In: IEEE International Conference on Acoustics, Speech and Signal Processing (ICASSP). 2023: 1-5.

[11] Lu B and Koji Ito. "Computation of multiple inverse kinematic solutions for redundant manipulators by inverting modular neural networks". IEEJ Transactions on Electronics, Information and Systems, 1995, 116(1): 49–56.

[12] Aaron D'Souza, Sethu Vijayakumar and Stefan Schaal. "Learning inverse kinematics". In: IEEE/RSJ International Conference on Intelligent Robots and Systems (IROS). 2001: 298–303.

[13] Eimei Oyama, Nak Young Chong, Arvin Agah et al. "Inverse kinematics learning by modular architecture neural networks with performance prediction networks". In: IEEE International Conference on Robotics and Automation (ICRA). 2001: 1006–1012.

[14] Kihyuk Sohn, Honglak Lee and Xinchen Yan. "Learning structured output representation using deep conditional generative models". In: Neural Information Processing Systems (NIPS). 2015: 3483–3491.

[15] Brian Ichter, James Harrison and Marco Pavone. "Learning sampling distributions for robot motion planning". In: IEEE International Conference on Robotics and Automation (ICRA). 2018: 7087–7094.

[16] Mehdi Mirza and Simon Osindero. "Conditional generative adversarial nets". arXiv preprint arXiv:1411.1784, 2014.

[17] Hailin Ren and Pinhas Ben-Tzvi. "Learning inverse kinematics and dynamics of a robotic manipulator using generative adversarial networks". Robotics and Autonomous Systems, 2020, 124: 103386.

[18] Teguh Santoso Lembono, Emmanuel Pignat, Julius Jankowski et al. "Learning constrained distributions of robot configurations with generative adversarial network". IEEE Robotics and Automation Letters, 2021, 6(2): 4233–4240.

[19] Lynton Ardizzone, Carsten Lüth, Jakob Kruse et al. "Guided image generation with conditional invertible neural networks". arXiv preprint arXiv:1907.02392, 2019.

[20] Jakob Kruse, Lynton Ardizzone, Carsten Rother et al. "Benchmarking invertible architectures on inverse problems". In: ICML Workshop on Invertible Neural Networks and Normalizing Flows. 2019.

[21] Qu W, Liu T, Wu X, et al. "Embodied Self-Supervised Learning (EMSSL) with Sampling and Training Coordination for Robot Arm Inverse Kinematics Model Learning". In: IEEE International Conference on Development and Learning (ICDL). 2023: 100-106.

[22] Ian Goodfellow. "Tutorial: Generative adversarial networks". In: Neural Information Processing Systems (NIPS). 2016.

[23] Thomas G Dietterich. "Ensemble methods in machine learning". In: International Workshop on Multiple Classifier Systems. 2000: 1–15.

[24] Zhou Z. "Integrated learning: fundamentals and algorithms". Beijing: Electronic Industry Press, 2020 (in Chinese).

[25] Qu W, Liu T, Luo D. "High-Precise Robot Arm Manipulation based on Online Iterative Learning and Forward Simulation with Positioning Error Below End-Effector Physical Minimum Displacement". arXiv preprint arXiv:2302.13338, 2023.